\documentclass[twocolumn]{bmcart}

\usepackage[utf8]{inputenc} %unicode support
\usepackage{xcolor}
\usepackage{cite}
\usepackage[sort&compress]{natbib}
\usepackage{balance}
\usepackage{amsmath,amssymb,amsfonts}
\usepackage{algorithmic}
\usepackage{graphicx}
\usepackage{textcomp}
\usepackage{siunitx}
\usepackage{multirow,tabularx}

\startlocaldefs
\endlocaldefs

\begin{document}
	
	%%% Start of article front matter
	\begin{frontmatter}
		
		\begin{fmbox}
			\dochead{Research}
			
			
			\title{Facial UV Map Completion for Pose-invariant Face Recognition: A Novel Adversarial Approach based on Coupled Attention Residual UNets}
			
			%%%%%%%%%%%%%%%%%%%%%%%%%%%%%%%%%%%%%%%%%%%%%%
			%%                                          %%
			%% Enter the authors here                   %%
			%%                                          %%
			%% Specify information, if available,       %%
			%% in the form:                             %%
			%%   <key>={<id1>,<id2>}                    %%
			%%   <key>=                                 %%
			%% Comment or delete the keys which are     %%
			%% not used. Repeat \author command as much %%
			%% as required.                             %%
			%%                                          %%
			%%%%%%%%%%%%%%%%%%%%%%%%%%%%%%%%%%%%%%%%%%%%%%
			
			\author[
			addressref={aff1},
			noteref={n1}, 
			email={ypencil@hanmail.net}
			]{\inits{}\fnm{In Seop} \snm{Na}}
			\author[
			addressref={aff2},                   % id's of 
			noteref={n1}, 
			email={bktranquangchung@gmail.com}   % email address
			]{\inits{Q}\fnm{Chung} \snm{Tran}}
			\author[
			addressref={aff3},
			email={nddung@ioit.ac.vn}
			]{\inits{D}\fnm{Dung} \snm{Nguyen}}
			\author[
			addressref={aff2},                   % id's of addresses, e.g. {aff1,aff2}
			corref={aff2},                       % id of  notes, if any
			email={sangdv@soict.hust.edu.vn}   % email address
			]{\inits{V}\fnm{Sang} \snm{Dinh}}

			%%%%%%%%%%%%%%%%%%%%%%%%%%%%%%%%%%%%%%%%%%%%%%
			%%                                          %%
			%% Enter the authors' addresses here        %%
			%%                                          %%
			%% Repeat \address commands as much as      %%
			%% required.                                %%
			%%                                          %%
			%%%%%%%%%%%%%%%%%%%%%%%%%%%%%%%%%%%%%%%%%%%%%%
			
			\address[id=aff1]{%
				\orgname{Chosun University, South Korea},
				\street{309 Pilmun-daero},
				\postcode{}
				\city{Gwangju 61452},
				\cny{South Korea}
			}
			\address[id=aff2]{%                           % unique id
				\orgname{Hanoi University of Science and Technology, Vietnam}, % university, etc
				\street{1 Dai Co Viet},                     %
				%\postcode{}                                % post or zip code
				\city{Hanoi},                              % city
				\cny{Vietnam}                                    % country
			}
			\address[id=aff3]{%
				\orgname{Vietnam Academy of Science and Technology},
				\street{18 Hoang Quoc Viet},
				\postcode{}
				\city{Hanoi},
				\cny{Vietnam}
			}
			
			%%%%%%%%%%%%%%%%%%%%%%%%%%%%%%%%%%%%%%%%%%%%%%
			%%                                          %%
			%% Enter short notes here                   %%
			%%                                          %%
			%% Short notes will be after addresses      %%
			%% on first page.                           %%
			%%                                          %%
			%%%%%%%%%%%%%%%%%%%%%%%%%%%%%%%%%%%%%%%%%%%%%%
			
			\begin{artnotes}
				%\note{Sample of title note}     % note to the article
				\note[id=n1]{Equal contributor} % note, connected to author
			\end{artnotes}
			
			%\end{fmbox}% comment this for two-column layout
			
			%%%%%%%%%%%%%%%%%%%%%%%%%%%%%%%%%%%%%%%%%%%%%%
			%%                                          %%
			%% The Abstract begins here                 %%
			%%                                          %%
			%% Please refer to the Instructions for     %%
			%% authors on http://www.biomedcentral.com  %%
			%% and include the section headings         %%
			%% accordingly for your article type.       %%
			%%                                          %%
			%%%%%%%%%%%%%%%%%%%%%%%%%%%%%%%%%%%%%%%%%%%%%%
			
			\begin{abstractbox}
				
				\begin{abstract} % abstract
					Pose-invariant face recognition refers to the problem of identifying or verifying a person by analyzing face images captured from different poses. This problem is challenging due to the large variation of pose, illumination and facial expression. A promising approach to deal with pose variation is to fulfill incomplete UV maps extracted from in-the-wild faces, then attach the completed UV map to a fitted 3D mesh and finally generate different 2D faces of arbitrary poses. The synthesized faces increase the pose variation for training deep face recognition models and reduce the pose discrepancy during the testing phase. In this paper, we propose a novel generative model called Attention ResCUNet-GAN to improve the UV map completion. We enhance the original UV-GAN by using a couple of U-Nets. Particularly, the skip connections within each U-Net are boosted by attention gates. Meanwhile, the features from two U-Nets are fused with trainable scalar weights. The experiments on the popular benchmarks, including Multi-PIE, LFW, CPLWF and CFP datasets, show that the proposed method yields superior performance compared to other existing methods.
				\end{abstract}
				
				%%%%%%%%%%%%%%%%%%%%%%%%%%%%%%%%%%%%%%%%%%%%%%
				%%                                          %%
				%% The keywords begin here                  %%
				%%                                          %%
				%% Put each keyword in separate \kwd{}.     %%
				%%                                          %%
				%%%%%%%%%%%%%%%%%%%%%%%%%%%%%%%%%%%%%%%%%%%%%%

			\end{abstractbox}
		\end{fmbox}% uncomment this for twcolumn layout
		
	\end{frontmatter}

	\section{Introduction}
	Face recognition has gained much attention for decades \cite{masi2018deep, zhou20183d, jafri2009survey}. Contrary to other popular biometrics, face recognition can be applied to uncooperative subjects in a non-instructive manner. While (near)-frontal face recognition has gradually matured, face recognition in the wild is still challenging due to different unconstrained factors. In fact, the performance of a face recognition system heavily depends on the pose of input faces. Recent studies show that the performance of face verification with the same view, such as frontal-frontal or profile-profile, is really good. However, the performance dramatically degrades when verifying faces in different views like frontal-profile \cite{tran2017disentangled}.
	
	Pose-invariant face recognition refers to the problem of identifying or verifying a person by analyzing face images captured from different poses. In recent years, numerous pose-invariant face recognition methods have been proposed. In \cite{parkhi2015deep, schroff2015facenet, sun2014deep, taigman2014deepface, yang2015weakly, sayan2020multimodal, sang2017facial, haiduong2019facial, blanco2015time}, the authors train deep neural networks on large-scale datasets to ease the effect of pose variation, which leads to significant improvements in the performance of face recognition. In \cite{masi2016we}, Masi et al. propose a method to enrich the pose variation in the training dataset by rotating faces across 3D space. Beyond, in \cite{sagonas2015robust}, Sagonas et al. propose a novel method to jointly learn both frontal view reconstruction and landmark localization by solving a constrained optimization problem. Kan et al. \cite{kan2014stacked} introduce stacked progressive auto-encoders (SPAE), which can learn pose-robust features through a complicated deep neural network to transform profile faces to frontal ones. In \cite{hassner2015effective}, Hassner et al. introduce a straightforward approach to generate frontal faces from a simple 3D shape. Peng et al. \cite{peng2017reconstruction} propose a new reconstruction loss for disentangled learning that encourages identity features of the same subject to be clustered together despite the pose variation. 
	
	Recently, generative adversarial networks (GANs) \cite{chongxuan2017triple} have proved to be powerful to mimic data distribution. GANs have been successfully applied to many computer vision tasks such as image inpainting \cite{peng2017reconstruction, yeh2016semantic, yu2018generative}, style transfer \cite{luan2017deep, zhu2017unpaired}, image synthesis \cite{karras2017progressive, karras2019style}, super-resolution \cite{ledig2017photo} and so on. These successful applications have motivated researchers to apply GANs to pose-invariant feature disentanglement \cite{tran2017disentangled, tran2018representation}, face completion \cite{wang2020recurrent} and face frontalization \cite{tran2017disentangled, huang2017beyond, yin2017towards, zhao2018towards, duan2020look}. In \cite{wang2020recurrent}, Wang et al. propose a recurrent generative adversarial network (RGAN), which consists of a CompletionNet and a DiscriminationNet, for completing face and recovering the missing region automatically. Dual et al. \cite{duan2020look} propose a boosting GAN (BoostGAN) for face deocclusion and frontalization. BoostGAN can generate photorealistic frontal faces with identity preservation from occluded but profile ones. TP-GAN \cite{huang2017beyond} uses a two-pathway GAN that simultaneously learns global structures and local information for photorealistic frontal view synthesis. Zhao et al. \cite{zhao2020recognizing} propose a unified deep architecture containing a face frontalization module and a discriminative learning module, which can be jointly learned in an end-to-end fashion. Zhang et al. \cite{zhang2020geometry} propose a geometry guided GAN to generate facial images with arbitrary expressions and poses conditioned on a set of facial landmarks. They embed a classifier into the GAN to facilitate image synthesis and perform facial expression recognition. In \cite{tran2018representation}, Tran et al. propose DR-GAN that can take one or multiple input images and produce one unified identity representation along with synthesized identity-preserved faces of various target poses. However, all methods mentioned above usually require a large amount of paired faces across different poses for training, which is overdemanding in real-world applications.
	
	\begin{figure*}[ht!]
		\includegraphics[width=350pt]{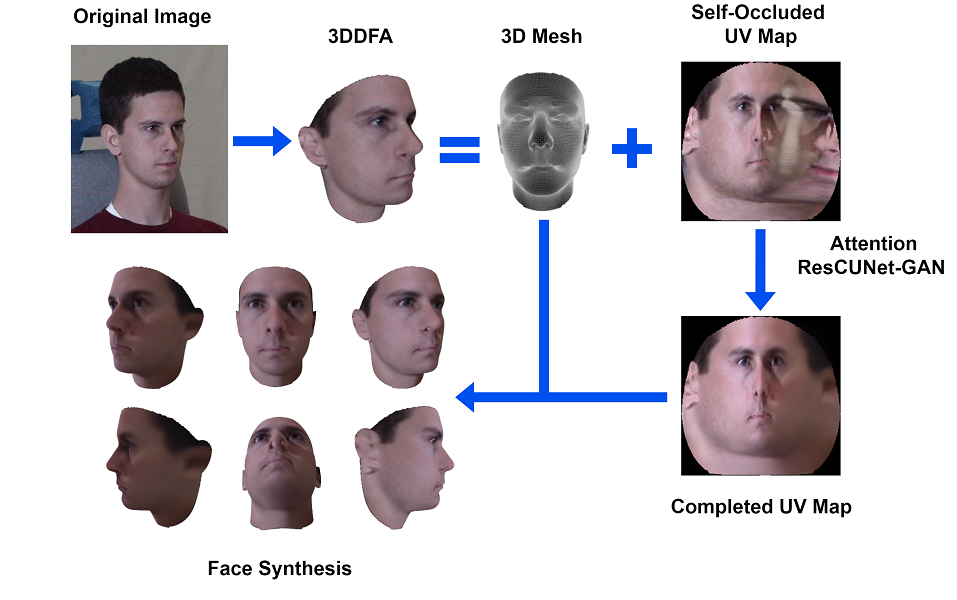}
		\caption{\textbf{A pipeline process of face synthesis.} Using 3DDFA to obtain a 3D mesh and an incomplete UV map. Then a new generative model is applied to recover the self-occluded regions. The completed UV map is attached to the fitted 3D mesh to generate faces of arbitrary poses.}
		\label{fig_pipeline}
	\end{figure*}
	
	In \cite{deng2018uv}, Deng et al. propose an adversarial UV map completion framework called UV-GAN to solve pose-invariant face recognition without the need of extensive pose coverage in the training dataset. The authors in \cite{deng2018uv} first fit a 3DMM \cite{booth20173d} to 2D profile face and get an incomplete UV map, which is then fulfilled by a straightforward pix2pix \cite{isola2017image, pix2pix2017}. The generator architecture in pix2pix follows the general shape of U-Net \cite{ronneberger2015u} to add skip connections between encoder and decoder subnetworks in order to enhance the transfer of low-level information between input and output. One weakness of the original UV-GAN is the plain architecture of the generator, which is shown to be worse than residual networks \cite{he2016deep}. Another weakness is that one U-Net block seems to be not enough to mix well low-level information in the encoder with high-level semantic features in the decoder. In \cite{xue2019side}, Deng et al. use UV-GAN with similar architecture as in \cite{deng2018uv} to extract side information as well as subspaces, and combine UV-GAN with robust PCA for the face recognition task.  He et al. \cite{he2019adversarial} introduce a framework for heterogeneous face synthesis from near-infrared (NIR) to visible domain. The framework consists of two adversarial generators to estimate a UV map and a facial texture map from an input NIR face, and then generate a corresponding frontal visible face. Nevertheless, both generators in this framework are based on the general U-Net structure \cite{ronneberger2015u, zhu2017unpaired}. Some efforts \cite{shah2018stacked, newell2016stacked} stack multiple U-Nets together, but skip connections are utilized only inside each single U-Net. Ibtehaz et al. \cite{ibtehaz2020multiresunet} propose residual paths with additional convolutional layers in skip connections to reduce the semantic gap between encoder and decoder features.  In \cite{schlemper2019attention}, Oktay et al. introduce attention gates to implicitly learn to suppress irrelevant regions in an input image while highlighting salient features useful for a specific task. In \cite{tang2019cu}, Tang et al. introduce coupled U-Nets architecture, where coupling connections are utilized to improve the information flow across U-Nets.  
	
	In this paper, we propose a new generative model architecture called Attention ResCUNet-GAN, where the generator is coupled U-Nets, and the backbone of each encoder is enhanced by residual network architecture. We use attention gates for skip connections within each U-Net to suppress irrelevant low-level information from encoders. We also use skip connections across two U-Nets to limit gradient vanishing and promote feature reuse. The experiments on the popular benchmarks demonstrate that our Attention ResCUNet-GAN yields considerably better results than the original UV-GAN model.
	
	The rest of this paper is organized as follows. Details of our proposed method are presented in Section 2. Section 3 presents our experimental results on the Multi-PIE dataset. Finally, the conclusion is made in Section 4.
	
	\section{Our Proposed Method}
	Following \cite{deng2018uv}, we use 3DDFA \cite{zhu2017face} to fitting 2D images to retrieve UV maps and 3D meshes. With a non-frontal face, the UV map generated by 3DDFA is always incomplete due to self-occlusion. Hence, we propose a new generative model architecture called Attention ResCUNet-GAN to improve the performance of the original UV-GAN \cite{deng2018uv} in filling up the missing contents of the UV map, which in turns helps to synthesize facial images of arbitrary poses. The overall pipeline process to synthesize more faces of various poses is depicted in Fig.~\ref{fig_pipeline}.
	
	\subsection{3DDFA Fitting}
	\subsubsection{3D Morphable Model} Blanz and Vetter \cite{blanz1999morphable} introduce the 3D Morphable Model Model (3DMM) to recover the 3D face from a 2D image. Assuming that a 3D face scan with $N$ vertexes can be represented as a $3N \times 1$ vector $\textbf{S} = [x_1, y_1, z_1, \ldots, x_N, y_N, z_N]^T \in \in\mathbb{R}^{3N}$, where $[x_i, y_i, z_i]^T$ are the object-centered Cartesian coordinates of the $i$-th vertex. Given a dataset of such 3D face scans, one would like to represent them as a smaller set of variables. The authors in \cite{blanz1999morphable} propose to use a two-stage principle component analysis (PCA) to estimate the shape identity parameters along with expression parameters of the 3D faces. Suppose that, after the first stage, we keep first $n_s$ principal components and $\textbf{s}_1, \textbf{s}_2, \ldots, \textbf{s}_{n_s}$ are the corresponding orthonormal basis, then a 3D face $\textbf{S}$ can be represented as follows: 
	\begin{align}
	\textbf{S} &= \bar{\textbf{S}} + \sum_{i=1}^{n_s}\textbf{s}_i\alpha_i
	%\hat{T} &= \bar{T} + \sum_{i=1}^{n_t}t_i\beta_i
	\end{align}
	where $\bar{\textbf{S}} \in\mathbb{R}^{3N}$ are the mean shape vector across the dataset of 3D face scans and $\boldsymbol\alpha = [\alpha_1, \ldots, \alpha_{n_s}]$  are the shape parameters.
	
	In the second stage, a new PCA is trained on the offsets between expression scans and neutral scans. After this stage, the final shape a representation is follows:
	\begin{align}
	\textbf{S} &= \bar{\textbf{S}} + \sum_{i=1}^{n_s}\textbf{s}_i\alpha_i + \sum_{i=1}^{n_e}\textbf{e}_i\beta_i, 
	%\hat{T} &= \bar{T} + \sum_{i=1}^{n_t}t_i\beta_i
	\end{align}
	where $\textbf{e}_i, i = 1,\ldots,n_e$ are the orthonormal basis of first $n_e$ principal components, and $\boldsymbol \beta = [\beta_1, \ldots, \beta_{n_e}]$ are the expression parameters. 
	
	After the 3D face is constructed, a rigid transformation is applied on the shape from the barycentric coordinate to camera based world coordinate. Each 3D vertex $\textbf{v} = [x, y, z]^T$ is rotated and translated as follows:
	\begin{equation}
	\textbf{v}_{cam} = \textbf{R}\textbf{v} + \textbf{t},
	\end{equation} 
	where $\textbf{R} \in \mathbb{R}^{3 \times 3}$ and $\textbf{t} = [t_x, t_y, t_z]^T $ are the 3D rotation and translation components, respectively.
	
	Finally, each 3D point can be projected into its 2D location in the image plane with scale orthographic projection:
	\begin{equation}
	\textbf{v}_p = f*\textbf{PR}*\textbf{v}_{cam} + \textbf{t}_{2d},
	\end{equation}
	where $f$ is the scale factor, $\textbf{Pr} =  \bigl(\begin{smallmatrix}1 & 0 & 0\\ 0 & 1 & 0\end{smallmatrix}\bigr)$ is the orthographic projection matrix and $\textbf{t}_{2d}$ is the principal point that is set to the image center.  
	
	Suppose that the set of all the model parameters are denoted by $\textbf{p} = [f, \textbf{R}, \textbf{t}_{2d}, \boldsymbol\alpha,  \boldsymbol\beta]$.
	
	\subsubsection{3DDFA method} 
	Method 3DDFA associates Cascaded Regression and a Convolutional Neural Network (CNN). Cascaded CNN can be formulated as:
	\begin{equation}
	\textbf{p}^{k+1} = \textbf{p}^k +{Net}^k(Feat(I,\textbf{p}^k)),
	\end{equation}
	where $\textbf{p}^k$ is the model parameters at the $k$-th iteration, which is updated by applying a CNN-based regressor ${Net}^k$ on the shape indexed feature $Feat$ that depends on the input image $\textbf{I}$ and the current parameters $\textbf{p}^k$. 
	
	The purpose of the CNN regressors is to predict the parameter update $\Delta \textbf{p}$ to shift the initial parameter $\textbf{p}^0$ as close as possible to the ground truth $\textbf{p}^{g}$. In term of objective function, \cite{zhu2017face} proposes to use the Optimized Weighted Parameter Distance Cost (OWPDC):
	\begin{align}
	E_{owpdc} = &(\Delta \textbf{p} + \textbf{p}^0 - \textbf{p}^{g})^T{\text{diag}}(\textbf{w}^*) \\ \nonumber 
	&(\Delta \textbf{p} + \textbf{p}^0 - \textbf{p}^{g}),
	\end{align}
	where $\textbf{w}^*$ is the optimized parameter importance vector.
	
	\subsection{Proposed Network for UV Map Completion}
	The proposed Attention ResCUNet-GAN consists of a generator, two discriminators, and an identity preserving module. The global discriminator deals with the global structure of entire complete UV maps, while the local discriminator focuses on the local details of the face region. 
	
	\subsubsection{Generator Network}
	An incomplete UV map is fed into Attention ResCUNet-GAN Generator, which acts as an auto-encoder to reconstruct missing regions. We use the following reconstruction loss as in \cite{deng2018uv}:
	\begin{equation}
	L_{rec} = \frac{1}{W * H} \sum_{i=1}^{W}\sum_{j=1}^{H}|G(I^P_{i,j})-I^F_{i,j}|,
	\end{equation}
	where $I^P$ is the input incomplete UV map, $G(I^P)$ is the output from the generator, and $I^F$ is the ground truth texture. 
	
	\begin{figure*} [ht!]
		\includegraphics[width=450pt]{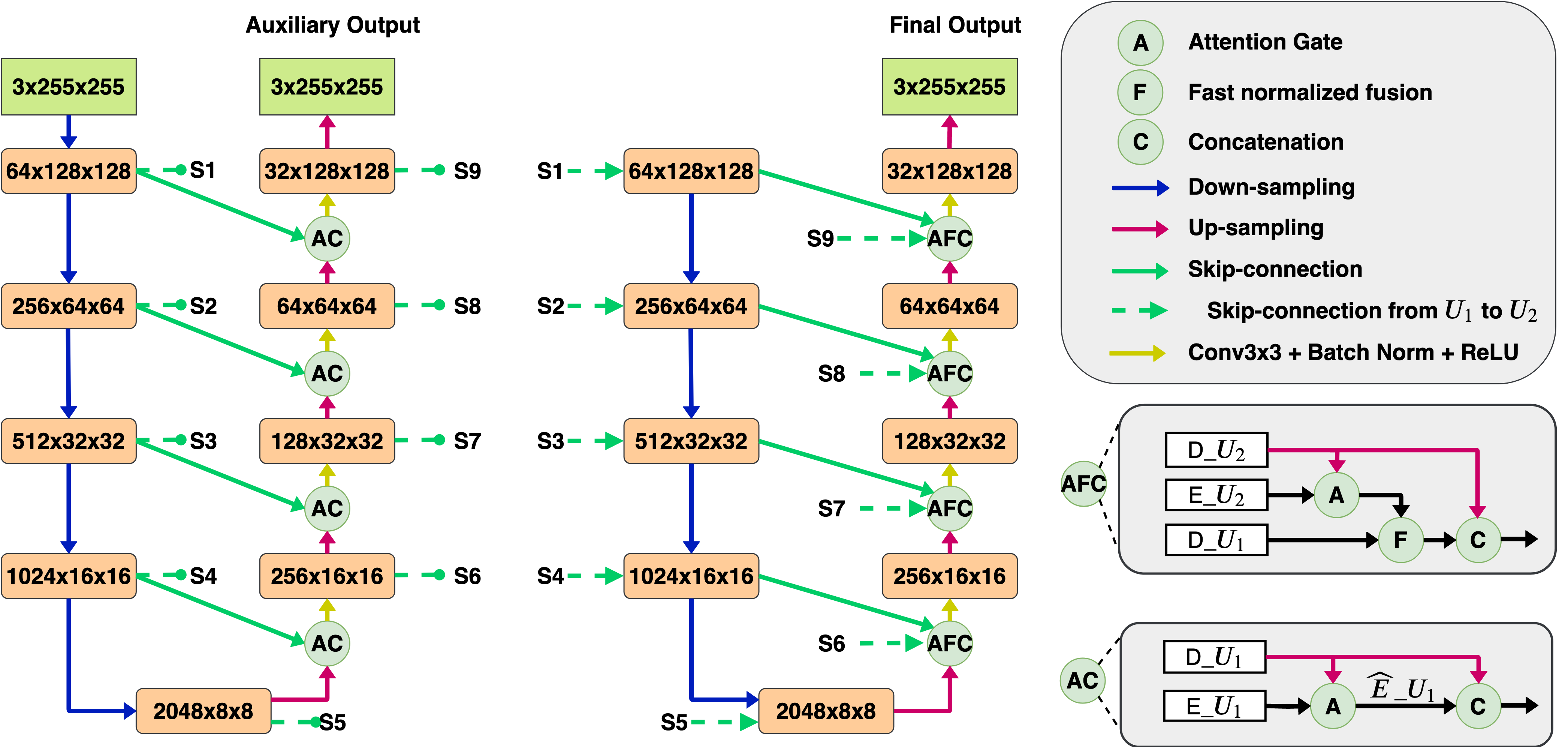}
		\caption{\textbf{Generator architecture.} The generator of proposed Attention ResCUNet-GAN consists of coupled U-Nets. Skip connections within each U-Net are enhanced with attention gates before concatenation. The contextual information from the first U-Net decoder is weighted fused with attentive low-level feature maps of the second U-Net encoder before concatenation with the high-level coarse feature maps of the second U-Net decoder. An auxiliary loss is used to improve gradient flow during the training phase.}
		\label{generator}
	\end{figure*}
	
	The generator (Fig.~\ref{generator}) consists of coupled U-Nets \cite{tang2019cu}. A drawback of the UV-GAN's generator is the plain convolutional backbone, which is shown to be rapidly degraded as the network depth increases \cite{he2016deep}. Therefore, here we leverage the residual architecture in \cite{he2016deep} to build a deeper backbone that is capable of extracting better high-level features without suffering from the degradation problem. Particularly, in terms of the backbone network for encoders, we use ResNet-50 \cite{he2016deep} consisting of multiple bottleneck residual blocks, each of which is a stack of three successive layers with 1x1, 3x3, 1x1 convolutions. Batch normalization is used right after each convolution and before activation layers. We use skip connections within each U-Net to transfer low-level information from the encoder to high-level contextual features in the decoder. Attention gates \cite{schlemper2019attention} are used to suppress irrelevant low-level information from encoders. Fig.~\ref{AG} illustrates how a coarse feature map can guide another low-level feature map to ignore irrelevant information.
	
	\begin{figure*} [ht!]
		\includegraphics[width=450pt]{Attention_Gates.png}
		\caption{\textbf{Attention gate (AG)}. The gating signal $g$ is obtained from a coarse feature map in the decoder, which provides information to disambiguate irrelevant information in the low-level feature map $x$ in the encoder. The concatenated features $x$ and $g$ are linearly mapped to a $F_i$-dimensional intermediate space. The attention mask $\theta$ guides the attention gate to capture only the important information $\hat{x}$.}
		\label{AG}
	\end{figure*}
	
	To combine features across two U-Nets, one can apply a direct depth-wise concatenation of the coarse feature maps $D\_U_1$, $D\_U_2$ extracted from the decoders of both U-Nets and the attentive information $\hat{E}\_U_2$ extracted from an attention gate of the encoder of the second U-Net. In such a combination, the latest feature map $D\_U_2$, which is thought to obtain more contextual information, would play the most crucial role regarding the contribution to the final output. However, such a direct concatenation always requires more memory. Thus, before concatenating with $D\_U_2$, here we apply fast normalized fusion \cite{tan2019efficientdet} to combine $D\_U_1$ and $\hat{E}\_U_2$ as follows:
	\begin{equation}
	\hat{D}\_U_1 = \frac{w_1 \times D\_U_1 + w_2 \times \hat{E}\_U_2}{w_1 + w_2 + \epsilon}
	\label{fastFusion}
	\end{equation}
	where $w_1, w_2$ are learnable scalar weights that can be trained via normal back propagation algorithm and $\epsilon = 0.0001$ is a small value to avoid numerical instability. Parameters   are ensured to be positive by applying Relu activation after them.
	
\begin{figure} 
		\includegraphics[width=0.45\textwidth]{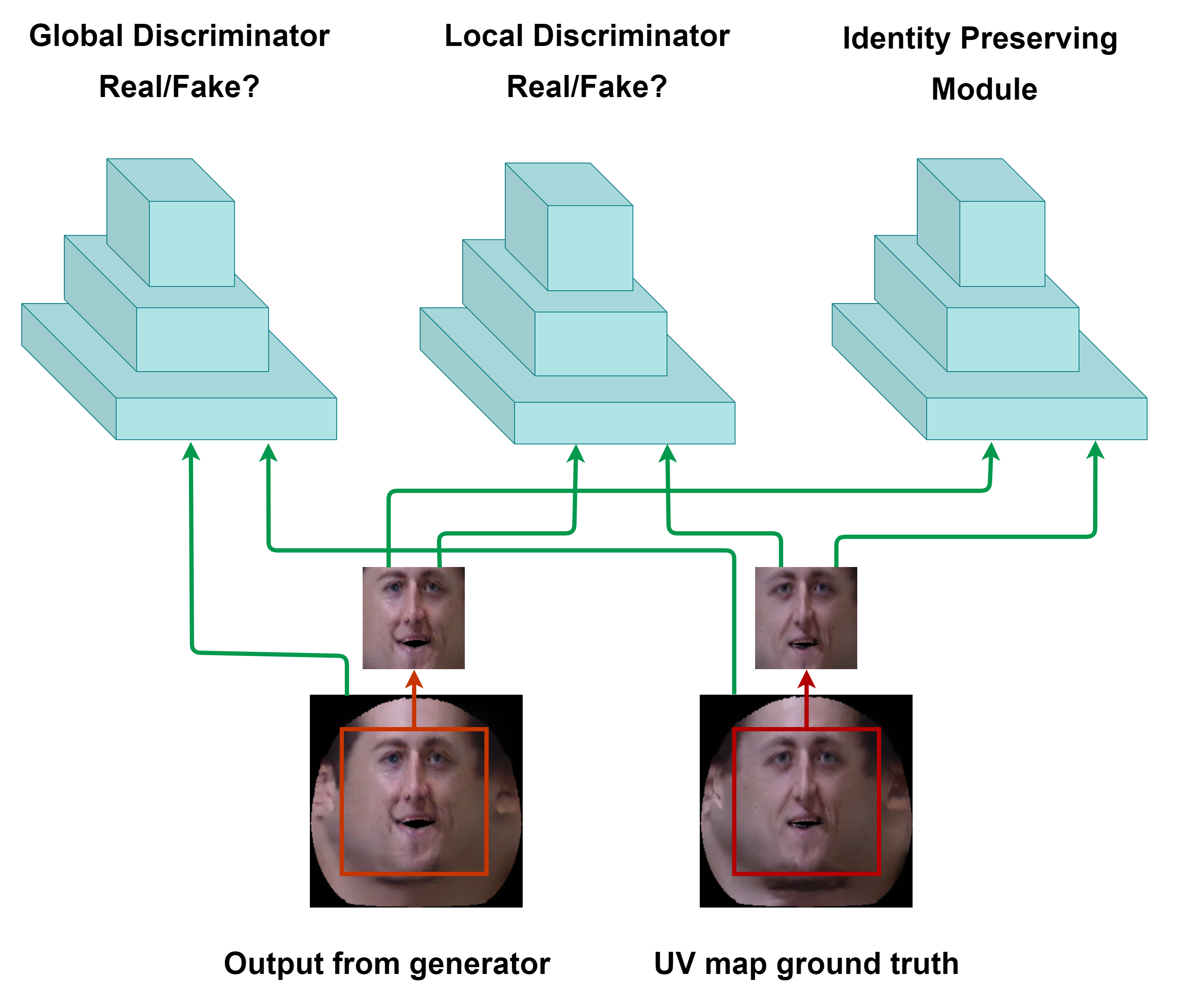}
		\caption{\textbf{Discriminators and identity preserving module of proposed Attention ResCUNet-GAN.} The global discriminator is responsible for the global structure of entire UV maps. The local discriminator focuses on the local facial details. The identity preserving module keeps the identity information unchanged during the modification of the generator.}
		\label{discriminator}
	\end{figure}

	\textbf{Global and Local Discriminators.} Global discriminator enforces maintaining the surrounding context of the facial image. Meanwhile, the local discriminator focuses on the central face region to enforce better recovering local details such as eye, nose, mouth and so on. We keep the same architectures for the discriminators as described in \cite{deng2018uv} (Fig.~\ref{discriminator}). The following typical adversarial loss is used:
	\begin{align}
	L_{adv} = & \min\limits_{G}\max\limits_{D} E_{x~p_d(x),y~p_d(y)}[log(D(x,y))] + \\ \nonumber
	&E_{x~p_d(x),z~p_d(z)}[log(1 -  D(G(x,z),y))], 
	\end{align}
	where $p_d(x), p_d(y), p_d(z)$ denote the distributions of incomplete UV maps $x$, complete UV maps $y$ and the Gaussian noise $z$, respectively.
	
	\textbf{Identity preserving module.} The synthetic faces must not only be photorealistic but also preserve identity information, which plays a crucial role in generation-based face recognition. To this end, the following identity loss \cite{deng2018uv} is used:
	\begin{equation}
	L_{id} = \parallel F(I^F) - F(G(I^p)) \parallel^2_2,
	\label{identity}
	\end{equation}
	where $F(.)$ denotes the embedding features extracted by the last layer before softmax in a pretrained CNN. Here in terms of embedding feature extractor, we use FaceNet pretrained on VGGFace2 dataset, which contains 3.31M face images of 9,131 identities. This feature extractor is frozen during training. The identity preserving module in Eq.~(\ref{identity}) enforces the embedding features of faces in the UV map ground truth $I^F$ and the generated UV map $G(I^P)$ to be close to each other. The dimension of the embedding features is 512.
	
	\textbf{Final loss function.} Overall, the total loss function is a weighted sum of the abovementioned losses:
	\begin{equation}
	L_{total} = L_{rec} + \lambda_1L^{local}_{adv} + \lambda_2L^{global}_{adv} + \lambda_3L_{id},
	\end{equation}
	where $\lambda_1, \lambda_2, \lambda_3$ are the weights that control the importance factors of different losses. 
	
	Moreover, a similar auxiliary loss is also applied to the intermediate output of the generator right after the end of the first U-Net decoder. The auxiliary loss strengthens the gradient flow to the layers of the first U-Net so that the parameters in the first U-Net can be trained more efficiently. Therefore, the final loss can be expressed as follows:
	\begin{equation}
	L_{final} = L_{total} + \eta L^{aux}_{total},
	\end{equation}
	where $\eta$ is a parameter regulating the contribution of the auxiliary loss.
	
	\section{Experiments and Evaluation}
	\subsection{Datasets and settings}
	We train our Attention ResCUNet-GAN on the Multi-PIE dataset \cite{gross2010multi}. All subjects in this dataset were taken in 15 viewpoints, 19 illumination conditions, and many facial expressions. Totally, there are more than 750,000 images of 337 people. 
	
\begin{figure}[ht!]
		\includegraphics[width=0.45\textwidth]{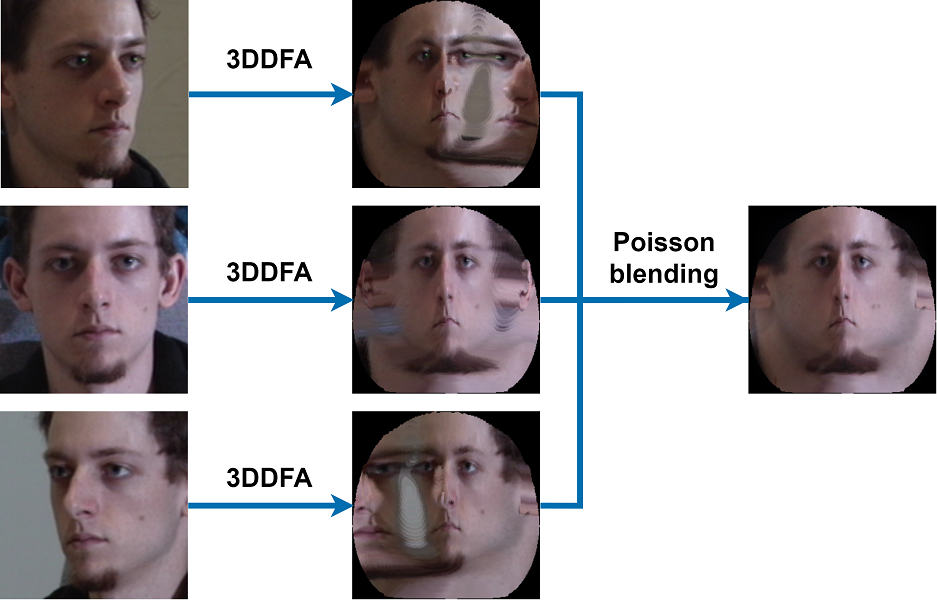}
		\caption{\textbf{The creation of ground-truth complete UV maps.} Three facial images with yaw angles of \ang{0}, \ang{-30}, \ang{+30} are fed to the 3DDFA model to create three incomplete UV maps which are then merged by Poisson blending to generate the ground-truth complete UV map.}
		\label{fig_poisson}
	\end{figure}

	For every subject with each illumination condition and facial expression, we feed 15 facial images captured from 15 viewpoints to the 3DDFA model to retrieve separate incomplete UV maps. We then select the incomplete UV maps with yaw angles of \ang{0}, \ang{-30}, \ang{+30} and merge them using Poisson blending \cite{perez2003poisson} (Fig.~\ref{fig_poisson}) to create the corresponding ground-truth UV map. In that way, we can ideally create 15 pairs of images for training the generator. Each of these pairs consists of an incomplete and a ground-truth UV map. However, in some cases, when the quality of an input facial image is not good enough, the 3DDFA model can not successfully detect the face landmarks; thus, the corresponding 3D mesh and incomplete UV map can not be created. Therefore, such cases are ignored in the training phase. All generated UV maps are rescaled to $256 \times 256$ to fit the input size of our ResCUNet-GAN. 
	
	\begin{figure*} [ht!]
		\includegraphics[width=450pt]{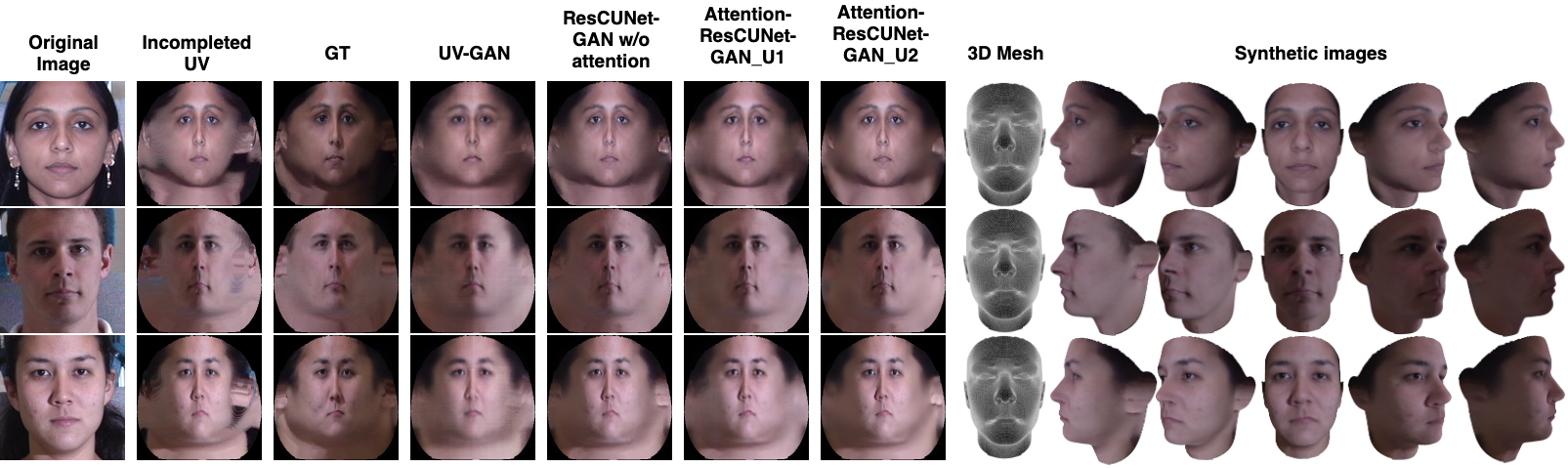}
		\caption{\textbf{Results with frontal input images.} Incomplete UV maps are generated using 3DDFA. Next columns are ground truth UV maps, results of UV-GAN, results of normal ResCUNet-GAN, intermediate results of Attention ResCUNet-GAN (after the first U-Net) and final results of Attention ResCUNet-GAN (after the second U-Net), respectively. The most right block shows some synthetic images generated based on the final results of Attention ResCUNet-GAN.}
		\label{frontal-result}
	\end{figure*}
	
	In addition to the proposed Attention ResCUNet-GAN, we also try a normal ResCUNet-GAN that has a similar architecture but without any attention gates and fast normalized fusion. In this ResCUNet-GAN, the concatenation is applied to all skip connections. Our networks are implemented in Pytorch. It takes three days for training each network on a server with two GPU RTX 2080Ti. We train each network for 100 epochs with a batch size of 16 and a learning rate of $10^{-4}$. We empirically set the importance factors as follows:  $\eta = 0.3, \lambda_1 = \lambda_2 = 0.5, \lambda_3 = 0.01$.
	
	In order to evaluate the effectiveness of the proposed method, we conduct experiments on pose-invariant face recognition on different benchmarks. Casia Web Face is a facial dataset that consists of 453,453 images over 10,575 identities. LFW (Labeled Faces in the Wild) is a well-known dataset for face verification in-the-wild. LFW contains more than 13,000 images of 1,680 identities, and each identity has two or more images of various poses. CPLFW (Cross-Pose LFW) is an extended version of LFW, which is more difficult due to different illuminations, occlusions, and expressions. CFP dataset consists of 500 subjects, each of which has ten frontal and four profile images. There are two evaluation protocols regarding the CFP dataset: frontal-frontal (FF) and frontal-profile (FP) face verification. Each of them has ten folders with 350 same-person pairs and 350 different-person pairs.

	\subsection{Image Reconstruction}
	
	We use two metrics to evaluate the quality of output from the Attention ResCUNet-GAN. The first metric is the structural similarity (SSIM), which is designed for measuring the similarity between images. The second one is the peak signal-to-noise ratio (PSNR), which is commonly used to measure the quality of reconstruction. Table~\ref{tab:eval-MultiPIE} shows that our method achieves better results than the original UV-GAN according to both metrics SSIM and PSNR. % The superior performance of our method is also confirmed by quantitative evaluations under different view changes shown in Table~\ref{tab:eval-MultiPIE-dif-poses}. 
	
	\begin{table}[ht!]
		\caption{Performance comparison of different methods on the Multi-PIE dataset}
		\begin{center}
			\begin{tabular}{ccc}
				\hline 
				\textbf{Model} & \textbf{SSIM} & \textbf{PSNR}  \\ 
				\hline \hline 
				UV-GAN \cite{deng2018uv} & 0.61 & 13.67 \\ \hline 
				Our ResCUNet-GAN (w/o attention gates & 0.66 & 15.67 \\ 
				and fast normalized fusion) && \\
				Our Attention ResCUNet-GAN & 0.685 & 15.974 \\
				\hline 
			\end{tabular} 
		\end{center}
		\label{tab:eval-MultiPIE}
	\end{table}

	%\begin{table*}[h!]
	%\caption{Quantitative evaluations under view changes}
	%\label{tab:eval-MultiPIE-dif-poses}
	%\begin{center}
	%\begin{tabular}{cccccccc}
	%\hline 
	%\textbf{Model} &
	%\textbf{Metric} & \textbf{\ang{0}} & \textbf{\ang{\pm15}} & \textbf{\ang{\pm30}} &
	%\textbf{\ang{\pm45}} &
	%\textbf{\ang{\pm60}} &
	%\textbf{\ang{\pm75}}  \\ 
	%\hline \hline 
	%\multirow{2}{*}{UV-GAN [3]}
	%&SSIM &0.619
	%&0.607&0.614
	%&0.611&0.603
	%&0.618 \\ 
	%&PSRN
	%&13.675&13.664
	%&13.683&13.669
	%&13.657&13.671 \\ \hline
	%\multirow{2}{*}{Our Attention ResCUNet-GAN}
	%&SSIM &0.687
	%&0.676&0.681
	%&0.682&0.697
	%&0.708 \\
	%&PSRN&16.22
	%&15.73&15.91
	%&15.85&16.23
	%&16.75 \\
	%\hline 
	%\end{tabular} 
	%\end{center}
	%\end{table*}
	
\begin{figure*} [ht!]
		\includegraphics[width=450pt]{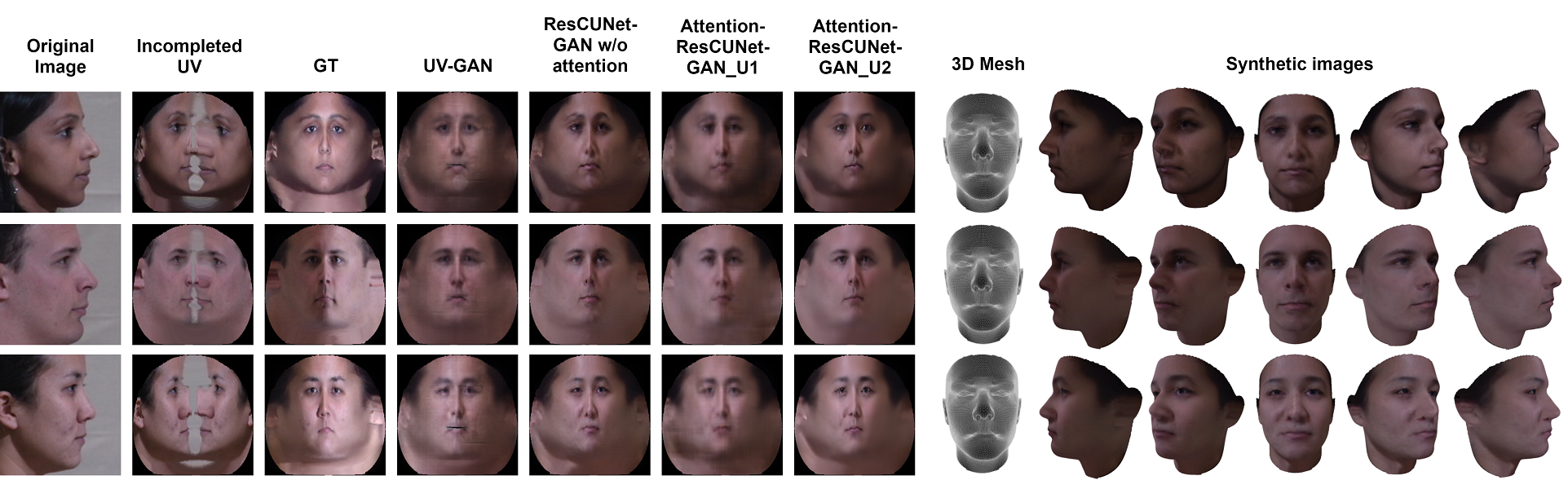}
		\caption{\textbf{Results with profile input images.} Incomplete UV maps are generated using 3DDFA. Next columns are ground truth UV maps, results of UV-GAN, results of normal ResCUNet-GAN, intermediate results of Attention ResCUNet-GAN (after the first U-Net) and final results of Attention ResCUNet-GAN (after the second U-Net), respectively. The most right block shows some synthetic images generated based on the final results of Attention ResCUNet-GAN.}
		\label{profile-result}
	\end{figure*}

\begin{figure*} [ht!]
		\includegraphics[width=450pt]{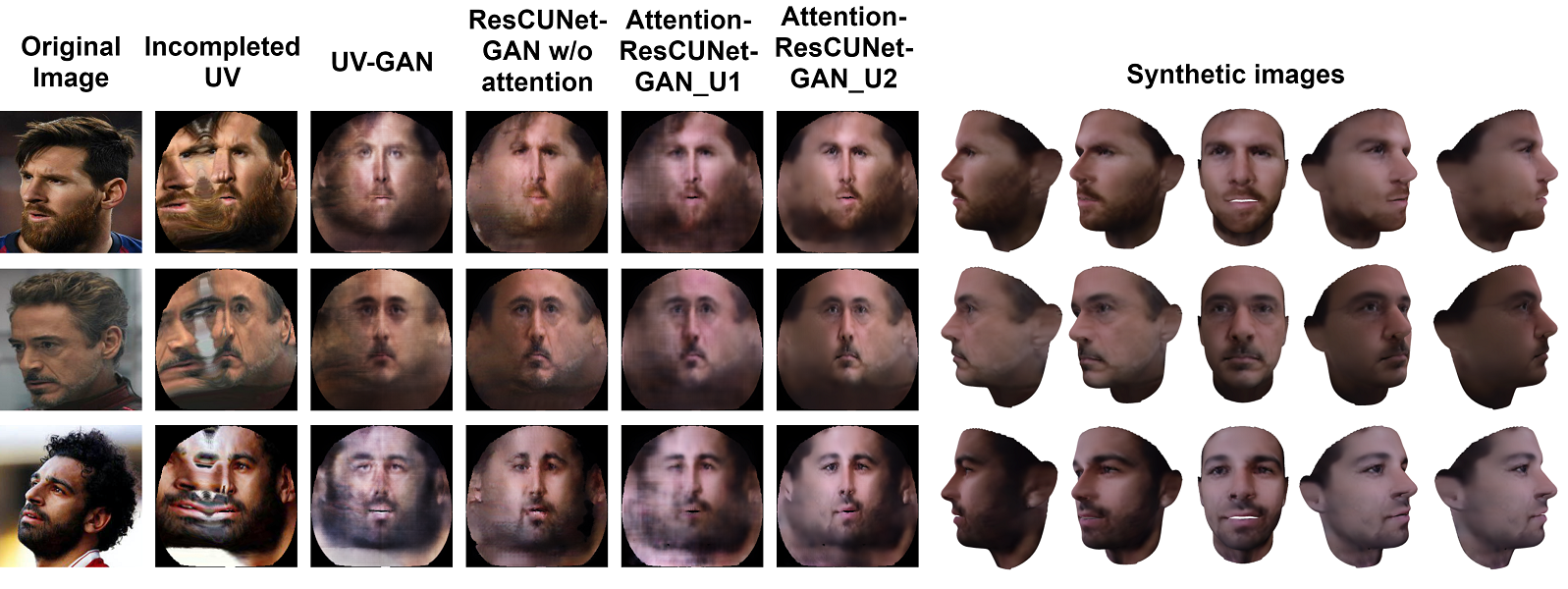}
		\caption{\textbf{Results with in-the-wild input images.} Incomplete UV maps are generated using 3DDFA. The ground truth UV maps are unavailable. The next columns are the results of UV-GAN, results of normal ResCUNet-GAN, intermediate results of Attention ResCUNet-GAN (after the first U-Net), and final results of Attention ResCUNet-GAN (after the second U-Net), respectively. The right block shows some synthetic images generated based on the final results of Attention ResCUNet-GAN.}
		\label{in-the-wild-result}
	\end{figure*}

	Fig.~\ref{frontal-result} and Fig.~\ref{profile-result} show the results of UV map completion on the test data taken from the Multi-PIE, where the UV map ground truths are available. For frontal input faces, the results of different methods look similar to each other. However, for profile input faces, the results are quite different. UV-GAN produces the worst UV maps. Normal ResCUNet-GAN yields better results, and Attention ResCUNet-GAN gives the most realistic ones with a smooth texture. Note that the intermediate output obtained from the first U-Net of the Attention ResCUNet-GAN still yields better results than UV-GAN's. 
	The results from some in-the-wild input images are shown in Fig.~\ref{in-the-wild-result}. One can see that Attention ResCUNet-GAN yields significantly better results than other ones, especially compared to the original UV-GAN.
	
\begin{figure*} [ht!]
		\includegraphics[width=450pt]{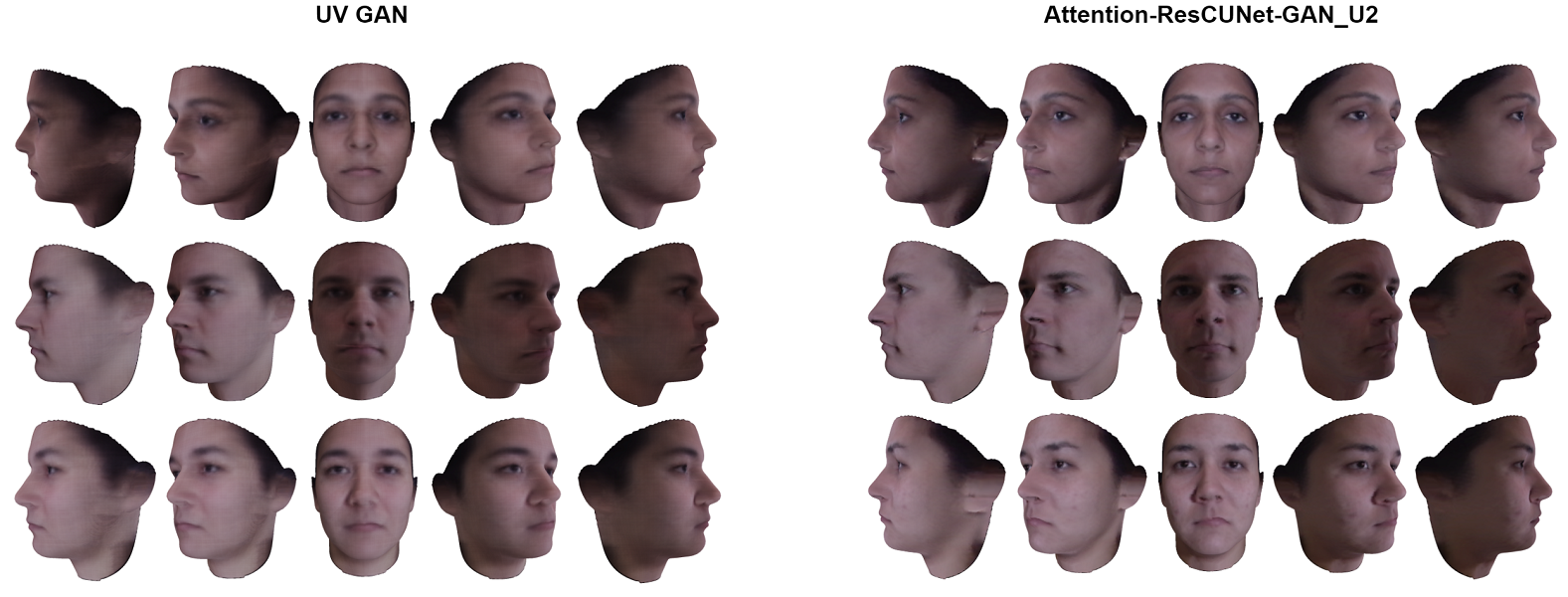}
		\caption{\textbf{Synthetic images for frontal input images.} The left block corresponds to the result of UV-GAN. The right block corresponds to the final result of Attention ResCUNet-GAN (after the second U-Net).}
		\label{frontal-compare}
	\end{figure*}

\begin{figure*} [ht!]
		\includegraphics[width=450pt]{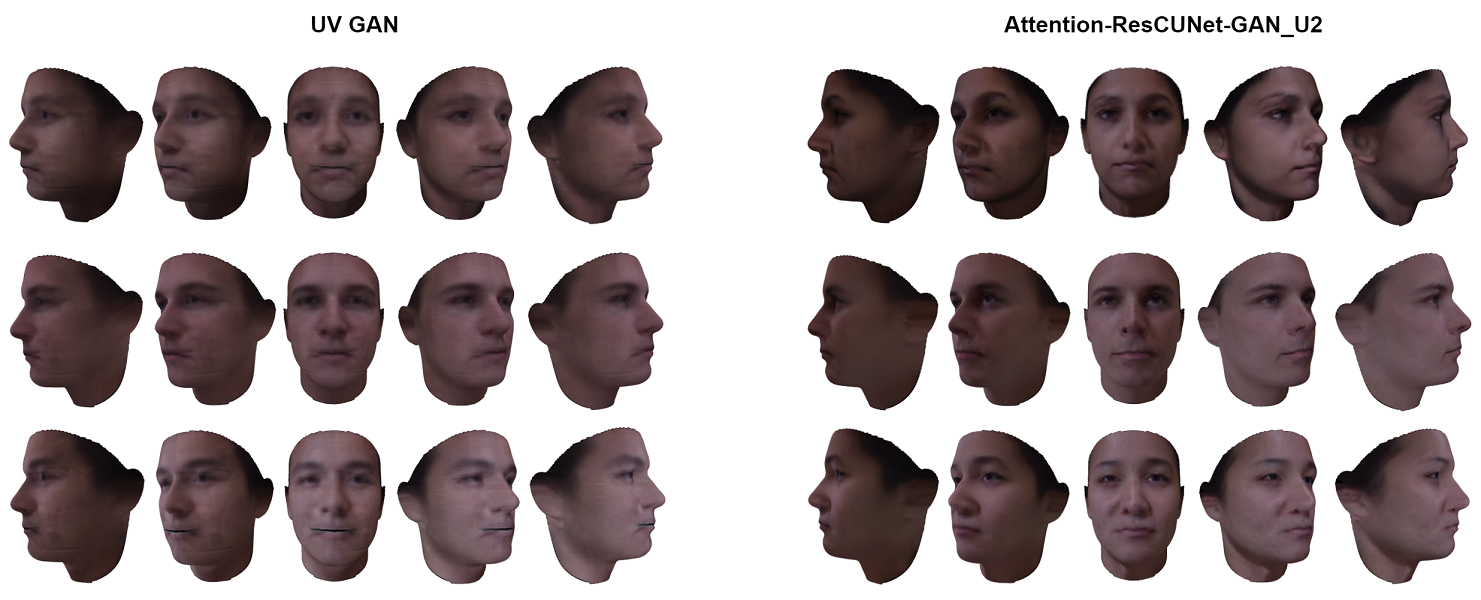}
		\caption{\textbf{Synthetic images for profile input images.} The left block corresponds to the result of UV-GAN. The right block corresponds to the final result of Attention ResCUNet-GAN (after the second U-Net).}
		\label{profile-compare}
	\end{figure*}

\begin{figure*} [ht!]
		\includegraphics[width=450pt]{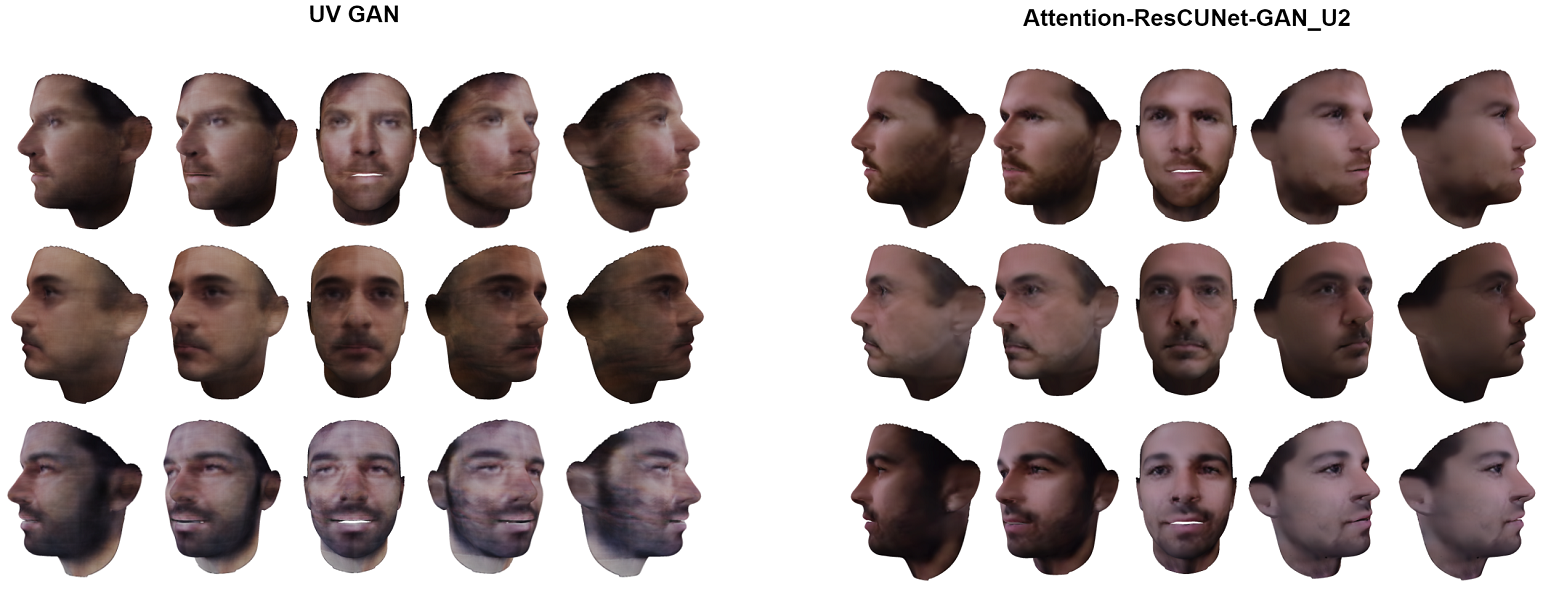}
		\caption{\textbf{Synthetic images for in-the-wild input images.} The left block corresponds to the result of UV-GAN. The right block corresponds to the final result of Attention ResCUNet-GAN (after the second U-Net).}
		\label{wild-compare}
	\end{figure*}

	In Fig.~\ref{frontal-compare}, Fig.~\ref{profile-compare} and Fig.~\ref{wild-compare}, we show side-by-side synthetic images generated from the UV map reconstructed by UV-GAN and the proposed Attention ResCUNet-GAN, respectively. One can see that our model yields qualitatively better results than the original UV-GAN, especially for profile and in-the-wild input images. 
	
\begin{figure}[ht!]
		\includegraphics[width=0.45\textwidth]{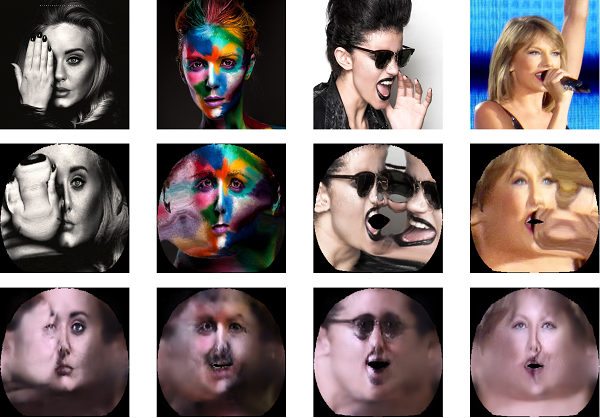}
		\caption{\textbf{Some failed cases when the input facial images are "abnormal" with respect to the training data.} The top row shows the input images, the second row contains incomplete UV map and the third row displays the completed UV maps generated by our Attention ResCUNet-GAN.}
		\label{fig_failure}
	\end{figure}

	The facial images in the Multi-PIE dataset are not diverse enough to reflect the real data distribution. Thus, in-the-wild faces occluded by strange things or with too much makeup can lead to some failures of the model, as illustrated in Fig.~\ref{fig_failure}.
	
	
	\subsection{Attention map visualization}
\begin{figure} [ht!]
		\includegraphics[scale=0.72]{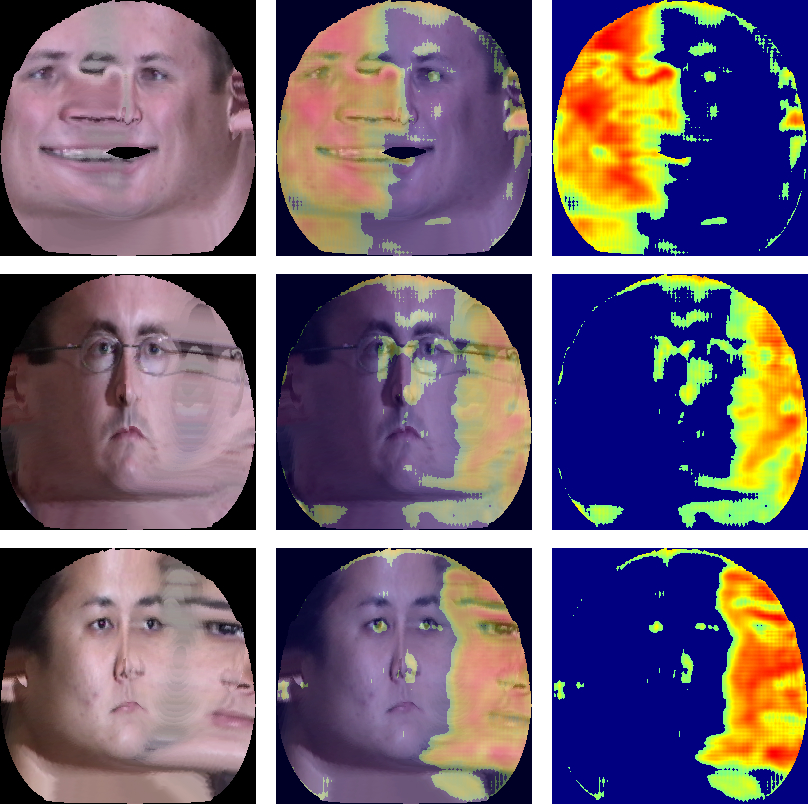}
		\caption{\textbf{Attention map visualization.} The first column contains UV maps generated by 3DDFA network, the second column contains generated UV maps overlaid by attention masks, and the last column illustrates attention coefficients only.}
		\label{visulization}
	\end{figure}

	The attention coefficients of the proposed Attention ResCUNet-GAN are visualized in Fig. \ref{visulization}. These attention coefficients are obtained in the attention gate of the AFC node that takes $S9$ as input (see Fig.~\ref{generator}). One can see that the attention maps try to ignore the visible face regions, focusing only on the missing regions of incomplete UV maps.
	
	\subsection{Pose Invariance Face Recognition}
	We compare our methods with UV-GAN on the Multi-PIE dataset in the face verification task. We take facial images from different pose ranged from \ang{0} to \ang{75} and frontalize them using UV-GAN and our methods. We then use a face detector \cite{7553523} to crop the central faces from the generated complete UV maps and push the cropped faces through ArcFace \cite{deng2019arcface} to verify if the synthetic frontal face and the ground truth one belong to the same subject or not. The verification results are shown on Table~\ref{ver-result}. One can see that the verification accuracy falls down along with the increase of pose. Nevertheless, our proposed ResCUNet-GANs (and even ResUNet-GAN with one U-Net block) always produces better frontal faces in term of preserving identity. Attention ResCUNet-GAN outperforms other methods by orders of magnitude on all profile poses. Surprisingly, for frontal faces, Attention ResCUNet-GAN yields little degraded results than normal ResCUNet-GAN. The reason may be that the useful information, which is necessary for the recognition task, in frontal images is almost comprehensive. Hence, a complicated transformer with attention gates and fast normalized fusion might unintendedly diminish some useful information and leads to the degradation in the verification accuracy. 
	
	\begin{table*}
		\caption{Verification results on different poses on the Multi-PIE dataset}
		\label{ver-result}
		\begin{center}
			\begin{tabular}{ccccccc}
				\hline 
				\textbf{Model} & \textbf{\ang{0}} & \textbf{\ang{\pm15}} & 
				\textbf{\ang{\pm30}} &
				\textbf{\ang{\pm45}} &
				\textbf{\ang{\pm60}} &
				\textbf{\ang{\pm75}}  \\ 
				\hline \hline 
				UV-GAN [3] &0.688&0.655&0.631&0.605&0.553&0.512 \\ 
				Our ResCUNet-GAN (w/o attention gates & 0.983&0.840&0.801&0.761&0.732&0.705 \\ 
				and fast normalized fusion) && \\
				Our Attention ResCUNet-GAN &0.955&0.874&0.837&0.794&0.768&0.742 \\
				\hline 
			\end{tabular} 
		\end{center}
	\end{table*}
	
	In the next experiment, we train a face recognition model on the CASIA dataset and evaluate its performance in the face verification task on other different datasets. Firstly, we train a face deep feature extractor with ResNet-101 backbone and arcface \cite{deng2019arcface} loss on the CASIA dataset augmented by using Attention ResCUNet-GAN. For each identity in CASIA, we generate different profile faces from the frontal one, ranging from \ang{-80} to \ang{80} with the step of \ang{20}. For each identity, we synthesize approximately 300 frontal and profile images. We train the network with a batch size of 128 for 30 epochs. The learned model is then used for the verification task on the LFW, CPLFW and CFP datasets. Note that for the CFP dataset, we consider two verification types: frontal-frontal means to verify two frontal faces, and frontal-profile means to verify a frontal face and a profile one.
	We use $k$-fold cross-validation to evaluate the face verification task. Particularly, each dataset will be divided into 10 groups ($k = 10$). Each group is considered as a test set in turn, while the remaining groups are used to tune the best verification threshold. In total, we have ten runs for each face verification dataset. The mean accuracy and the standard deviation over ten runs are reported.
	Table~\ref{tab:LWF-CPLFW} and Table~\ref{tab:CFP} show that data augmentation using the proposed Attention ResCUNet-GAN improves the performance of the recognition model. Note that the LFW dataset does not pay much attention to cross-pose face verification, and most faces in this dataset are nearly frontal. Therefore, a heavy facial pose augmentation using generative networks for training the recognition model is probably not really necessary. In fact, the verification performance on the LFW dataset over ten runs slightly fluctuates when we apply the proposed generative model for data augmentation. The standard deviation of accuracy increases from $0.032$ to $0.391$ (see Table~\ref{tab:LWF-CPLFW}). However, in overall, using Attention ResCUNet-GAN still helps to improve the average cross-validation accuracy. In contrast to the LFW dataset, the CPLFW dataset has lots of positive face pairs with different poses to enlarge intra-class variance. In this case, our model results in more stable improvements, where the standard deviation of accuracy is almost the same as if the data augmentation is not used. 
	\begin{table*}[ht!]
		\caption{Verification accuracy (\%) comparison on the
			LFW and CPLFW datasets}
		\label{tab:LWF-CPLFW}
		\begin{center}
			\begin{tabular}{ccc}
				\hline 
				\textbf{Model} & \textbf{LFW} & \textbf{CPLFW} \\ 
				\hline \hline 
				HUMAN-Individual &
				97.27&
				81.21\\
				CASIA-sm-augUVGAN \cite{deng2018uv} &
				99.22&
				-\\
				%CASIA-arcface-augUVGAN &
				%???&
				%???\\
				CASIA-arcface&
				99.43 $\pm$ 0.032&
				90.30 $\pm$ 1.797\\
				CASIA-arcface-augAttentionResCUNetGAN&
				99.47 $\pm$ 0.391&
				90.45 $\pm$ 1.812\\
				\hline 
			\end{tabular} 
		\end{center}
	\end{table*}     
	
\begin{figure} [ht!]
		\includegraphics[width=0.45\textwidth]{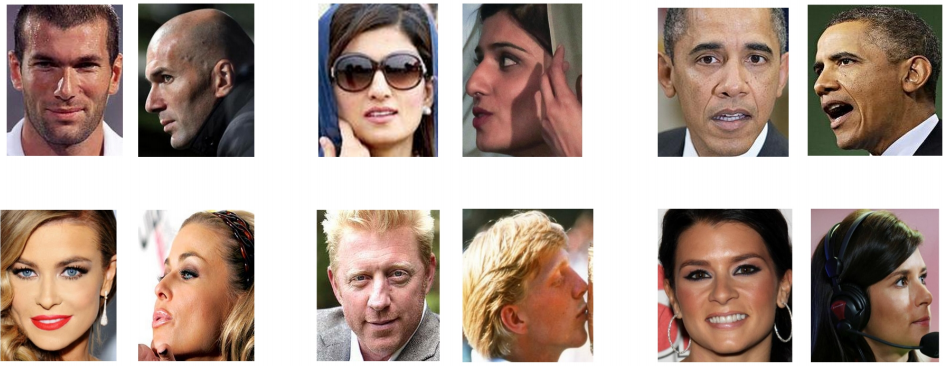}
		\caption{\textbf{Some samples of positive pairs from the CFP dataset.}
			\label{CFP}}
	\end{figure}

	The CFP dataset focuses on the pose variation in terms of extreme pose where many details of faces are occluded (see Fig.~\ref{CFP}). One can see from the Table~\ref{tab:CFP}, our Attention ResCUNetGAN considerably improves the performance of the face recognition model, especially for the frontal-profile subtask.
	
	\begin{table*} [ht!]
		\caption{Verification accuracy (\%) comparison on the CFP dataset}
		\label{tab:CFP}
		\begin{center}
			\begin{tabular}{ccc}
				\hline 
				\textbf{Model} & \textbf{Frontal-Frontal} & \textbf{Frontal-Profile} \\ 
				\hline \hline 
				CASIA-center \cite{deng2018uv} &
				98.34 $\pm$ 0.44 &
				87.77 $\pm$ 2.39
				\\ 
				CASIA-Sphere \cite{deng2018uv}&
				98.64 $\pm$ 0.24&
				84.39 $\pm$ 2.59 \\
				CASIA-sm \cite{deng2018uv}&
				98.59 $\pm$ 0.21 &
				87.74 $\pm$ 1.07 \\
				CASIA-sm-aug1 \cite{deng2018uv} &
				98.25 $\pm$ 0.42 &
				90.14 $\pm$ 1.53 \\ \hline
				CASIA-sm-augUV-GAN \cite{deng2018uv}&
				98.83 $\pm$ 0.27 &
				93.09 $\pm$ 1.72 \\
				-Profile2Frontal&
				-&
				93.55 $\pm$ 1.67\\
				-Frontal2Profile&
				-&
				93.72 $\pm$ 1.59\\
				-Template2Template&
				-&
				94.05 $\pm$ 1.73\\
				\hline \hline
				CASIA-arcface-augAttentionResCUNetGAN &
				99.47 $\pm$ 0.29 &
				97.0 $\pm$ 0.82 \\
				\hline 
			\end{tabular} 
		\end{center}
	\end{table*}
	
	\section{Conclusions and future work}
	In this paper, we introduce a novel generative model called Attention ResCUNet-GAN to generate complete facial UV maps, which allows us to synthesize various faces of arbitrary poses and improve pose-invariant face recognition performance. We leverage the residual connections in ResNet, intra-block and extra-block feature fusion in coupled UNets to enhance the generator. The skip connections within each U-Net are amplified with attention gates, while the contextual feature maps from two U-Nets are fused with trainable scalar weights. We jointly train global and local adversarial losses with identity preserving loss. The experiments show that the proposed Attention ResCUNet-GAN outperforms the original UV-GAN by order of magnitude in terms of both reconstruction metrics and the performance on the pose-invariant face verification task.
	
	In future work, we would like to exploit some recent efficient backbones such as EfficientNet \cite{tan2019efficientnet} to improve the performance of the proposed approach. More complex short-cut connections \cite{ibtehaz2020multiresunet, zhou2018unet} can also be utilized to improve gradient flow and stimulate feature reuse within the network.

	\begin{backmatter}
		
		\section*{Acknowledgements}
		The authors would like to thank Vietnam Artificial Intelligence System (VAIS) for providing computational resources to complete this work.     
		
		\section*{Author's contributions}
		
		In Seop Na: Formal analysis, Data curation, Validation, Funding acquisition, Investigation, Writing - review \& editing.  
		
		Chung Tran: Software, Data curation, Formal analysis, Visualization.
		
		Dung Nguyen: Investigation, Validation, Writing - review \& editing.
		
		Sang Dinh: Conceptualization, Methodology, Project administration, Investigation, Supervision, Resources, Writing - original draft. 
		
		\section*{Funding}
		This work was supported by the National Research Foundation of Korea (NRF) grant NRF-2019K2A9A1A06100184 and partially supported by the Vietnam Academy of Science and Technology under the grant number QTKR01.01/20-21. This work was also sponsored by the U.S. Army Combat Capabilities Development Command (CCDC) Pacific and CCDC Army Research Laboratory (ARL) under Contract Number W90GQZ-93290007. The views and conclusions contained in this document are those of the authors and should not be interpreted as representing the official policies, either expressed or implied, of the CCDC Pacific and CCDC ARL and the U.S. Government. The U.S. Government is authorized to reproduce and distribute reprints for Government purposes notwithstanding any copyright notation hereon. 
		
		\section*{Availability of data and materials}
		Not applicable
		
		\section*{Competing interests}
		The authors declare that they have no competing interests.
		
		\bibliographystyle{bmc-mathphys} % Style BST file (bmc-mathphys, vancouver, spbasic).
		\bibliography{bmc_article}      % Bibliography file (usually '*.bib' )

	\end{backmatter}
\end{document}